\documentclass[11pt,a4paper]{article}
\usepackage[hyperref]{naaclhlt2018}
\usepackage{times}
\usepackage{latexsym}
\usepackage{multirow}
\usepackage{graphicx}

\usepackage{url}

\aclfinalcopy 

\title{Annotating Student Talk in Text-based Classroom Discussions}

\author{Luca Lugini, Diane Litman, Amanda Godley, Christopher Olshefski\\
  University of Pittsburgh \\
  Pittsburgh, PA 15260 \\
  {\tt \{lul32,dlitman,agodley,cao48\}@pitt.edu}}

\date{}

\begin{document}
\maketitle
\begin{abstract}
Classroom discussions in English Language Arts have a positive effect on students' reading, writing, and reasoning skills.
Although prior work has largely focused on teacher talk and student-teacher interactions, 
we focus on three theoretically-motivated aspects of high-quality student talk: argumentation, specificity, and knowledge domain.
We introduce an annotation scheme, then show that the scheme can be used to produce reliable annotations and that the annotations are predictive of discussion quality.
We also highlight opportunities provided by our scheme for educational and natural language processing research.
\end{abstract}

\section{Introduction}
Current research, theory, and policy surrounding K-12 instruction in the United States highlight the role of student-centered disciplinary discussions
(i.e. discussions related to a specific academic discipline or school subject such as physics or English Language Arts)
in instructional quality and student learning opportunities \cite{Danielson:11,Grossman:14}. Such student-centered discussions -- often called ``dialogic" or ``inquiry-based'' -- are widely viewed as the most effective instructional approach for disciplinary understanding, problem-solving, and literacy \cite{Elizabeth:12,Engle:02, Murphy:09}. In English Language Arts (ELA) classrooms, student-centered discussions about literature have a positive impact on the development of students' reasoning, writing, and reading skills \cite{Applebee:03, Reznitskaya:13}. However, most studies have focused on the role of teachers and their talk \cite{Bloome:05,Elizabeth:12,Michaels:08} rather than on the aspects of student talk that contribute to discussion quality. 

Additionally, studies of student-centered discussions rarely use the same coding schemes, making it difficult to generalize across studies \cite{Elizabeth:12,Soter:08}. This limitation is partly due to the time-intensive work required to analyze discourse data through qualitative methods such as ethnography and discourse analysis. Thus, qualitative case studies have generated compelling theories about the specific features of student talk that lead to high-quality discussions, but few findings
can be generalized and leveraged to influence instructional improvements across ELA classrooms. 

As a first step towards developing an automated system for detecting the features of student talk that lead to high quality discussions, we propose a new annotation scheme for student talk during ELA ``text-based" discussions - that is, discussions that center on a text or piece of literature (e.g., book, play, or speech).
The annotation scheme was developed to capture three aspects of classroom talk that are theorized in the literature as important to discussion quality and learning opportunities: {\it argumentation} (the process of systematically reasoning in support of an idea),
{\it specificity} (the quality of belonging or relating uniquely to a particular subject),
and {\it knowledge domain} (area of expertise represented in the content of the talk).
We demonstrate the reliability and validity of our scheme via an annotation study of five transcripts of classroom discussion.


\section{Related Work}
One discourse feature used to assess the quality of discussions is students' argument moves: their claims about the text, their sharing of textual evidence for claims, and their warranting or reasoning to support the claims \cite{Reznitskaya:09,Toulmin:58}. Many researchers view student reasoning as of primary importance, particularly when the reasoning is elaborated and highly inferential \cite{Kim:14}. In Natural Language Processing (NLP), most educationally-oriented argumentation research 
has focused on corpora of student persuasive essays \cite{Ghosh:16,Klebanov:16,Persing:16,Wachsmuth:16,Stab:17,Nguyen:18}.
We instead focus on 
multi-party spoken discussion transcripts from classrooms. A second key difference consists in the inclusion of the warrant label in our scheme, as it is important to understand how students explicitly use reasoning to connect evidence to claims. 

Educational studies suggest that discussion quality is also influenced by the specificity of student talk \cite{Chisholm:11,Sohmer:09}. Chisholm and Godley found that as specificity increased, the quality of students' claims and reasoning also increased. Previous NLP research has studied specificity in the context of professionally written newspaper articles
\cite{Li:15,Li:16,Louis:11,Louis:12}.
While the annotation instructions used in these studies work well for general purpose corpora,
specificity in text-based discussions 
also needs to capture particular relations between discussions and texts. Furthermore, since the concept of a sentence is not clearly defined in speech,
we annotate argumentative discourse units rather than sentences  (see Section~\ref{sec:scheme}). 

The knowledge domain of student talk may also matter, that is, whether the talk focuses on disciplinary knowledge or lived experiences. Some research suggests that disciplinary learning opportunities are maximized when students draw on evidence and reasoning that are commonly accepted in the discipline \cite{Resnick:15}, although some studies suggest that evidence or reasoning from lived experiences increases discussion quality \cite{Beach:01}.
Previous related work in NLP analyzed evidence type for argumentative tweets \cite{addawood:16}. Although the categories of evidence type are different, their definition of evidence type is in line with our definition of knowledge domain. However, our research is distinct from this research in its application domain (i.e. social media vs. education) and in analyzing knowledge domain for all argumentative components, not only those containing claims.

\section{Annotation Scheme}
\label{sec:scheme}
Our annotation scheme\footnote{The  coding manual is in the supplemental material.} uses argument moves as the unit of analysis. We define an argument move as an utterance, or part of an utterance, that contains an argumentative discourse unit (ADU) \cite{Peldszus:13}.
Like Peldszus and Stede \shortcite{Peldszus:15}, in this paper we use  transcripts already segmented into argument moves and focus on the steps following segmentation, i.e., labeling argumentation, specificity, and knowledge domain.
Table \ref{tab:examples} shows a section of a transcribed classroom discussion along with labels assigned by a human annotator following segmentation.
\begin{table*}[ht]
\centering
\begin{tabular}{|c|c|p{0.425\linewidth}|c|c|c|}
\hline
\textbf{Move} & \textbf{Stu} & \textbf{Argument Move} & \textbf{Argument} & \textbf{Specificity} & \textbf{Domain}\\ \hline
23 & S1 & She's like really just protecting Willy from everything.  & claim & medium   & disciplinary  \\ \hline
24 & S1 & Like at the end of the book remember how she was telling the kids to leave and never come back. & evidence  & medium & disciplinary     \\ \hline
25 & S1 & Like she's not even caring about them, she's caring about Willy.  & warrant & medium & disciplinary   \\ \hline
\multicolumn{1}{c}{} \\[-0.15cm]
\hline
41 & S2 & It's like she's concerned with him trying to [inaudible] and he's concerned with trying to make her happy, you know? So he feels like he's failing when he's not making her happy like  & claim & high  & disciplinary \\ \hline
42 & S2 & "Let's bring your mother some good news"  & evidence  & high  & disciplinary\\ \hline
43 & S2 & but she knew that, there wasn't any good news,  so she wanted to act happy so he wouldn't be in pain. & warrant & high  & disciplinary \\ \hline
\multicolumn{1}{c}{} \\[-0.15cm]
\hline
55 & S3 & Some people they just ask for a job is just like, some money. & evidence  & low & experiential \\ \hline
\end{tabular}
\caption{Examples of argument moves and their respective annotations from a discussion of the book \textit{Death of a Salesman}. As shown by the argument move numbers, boxes for students S1, S2, and S3 indicate separate, non contiguous excerpts of the discussion.}
\label{tab:examples}
\end{table*}

\subsection{Argumentation}
The argumentation scheme is based on \cite{Lee:06} and consists of a simplified set of labels derived from Toulmin's \shortcite{Toulmin:58} model:
$(i)$ {\it Claim}: an arguable statement that presents a particular interpretation of a text or topic.
$(ii)$ {\it Evidence}: facts, documentation, text reference, or testimony used to support or justify a claim.
$(iii)$ {\it Warrant}: reasons explaining how a specific evidence instance supports a specific claim.
Our scheme specifies that warrants must come after claim and evidence, since by definition warrants cannot exist without them.

The first three moves in Table \ref{tab:examples} show a natural expression of an argument: a student first claims that Willy's wife is only trying to protect him, then provides a reference as evidence by mentioning something she said to her kids at the end of the book, and finally explains how
not caring
about her kids ties the evidence to the initial claim.
The second group shows the same argument progression, with evidence given as a direct quote.

\subsection{Specificity}
Specificity annotations are based on \cite{Chisholm:11} and have the goal of capturing text-related characteristics expressed in student talk.
Specificity labels are directly related to four distinct elements for an argument move:
(1) it is specific to one (or a few) character or scene;
(2) it makes significant qualifications or elaborations;
(3) it uses content-specific vocabulary (e.g. quotes from the text);
(4) it provides a chain of reasons.
Our annotation scheme for specificity includes three labels along a linear scale:
$(i)$ {\it Low}: statement that does not contain any of these elements.
$(ii)$ {\it Medium}: statement that accomplishes one of these elements.
$(iii)$ {\it High}: statement that clearly accomplishes at least two specificity elements.
Even though we do not explicitly use labels for the four specificity elements, we found that explicitly breaking down specificity into multiple components helped increase reliability when training annotators.

The first three argument moves in Table \ref{tab:examples} all contain the first element, as they refer to select characters in the book. However, no content-specific vocabulary,  clear chain of reasoning, or significant qualifications are provided; therefore all three moves are labeled as medium specificity.
The fourth move, however, accomplishes the first and fourth specificity elements, and is labeled as  high specificity.
The fifth move is also labeled high specificity since it is specific to one character/scene, and provides a direct quote from the text.
The last  move is labeled as low specificity as it reflects an overgeneralization about all humans.

\subsection{Knowledge Domain}

The possible labels for knowledge domain are:
$(i)$ {\it Disciplinary}: the statement is grounded in knowledge gathered from a text (either the one under discussion or others), such as a quote or a description of a character/event. 
$(ii)$ {\it Experiential}: the statement is drawn from human experience, such as what the speaker has experienced or thinks that other humans have experienced.

In Table \ref{tab:examples} the first six argument moves are labeled as disciplinary, since the moves reflect knowledge from the text currently being discussed. The last move, however, draws from a student's experience or perceived knowledge about the real world.

\section{Reliability and Validity Analyses}
\label{sec:data}
We carried out a reliability study for the proposed scheme using two pairs of expert annotators, P1 and P2. The annotators were trained by coding one transcript at a time and discussing disagreements.
Five text-based discussions were used for testing reliability after training:
pair P1 annotated discussions of \textit{The Bluest Eye}, \textit{Death of a Salesman}, and \textit{Macbeth}, while pair P2 annotated two separate discussions of \textit{Ain't I a Woman}.
250 argument moves (discussed by over 40 students and consisting of over 8200 words) were annotated. Inter-rater reliability was assessed using Cohen's kappa: unweighted  for argumentation and knowledge domain, but quadratic-weighted
 for specificity given its ordered labels.

\begin{table}[t]
\centering
\begin{tabular}{|c|c|c|c|}
\hline
\textbf{Moves} & \textbf{\begin{tabular}[c]{@{}c@{}}Argumen-\\ tation\\ (kappa)\end{tabular}} & \textbf{\begin{tabular}[c]{@{}c@{}}Specificity\\ (qwkappa)\end{tabular}} & \textbf{\begin{tabular}[c]{@{}c@{}}Domain\\ (kappa)\end{tabular}} \\ \hline
169            & 0.729                                                                        & 0.874                                                                    & 0.980                                                             \\ \hline
81             & 0.725                                                                        & 0.930                                                                    & 1                                                                 \\ \hline
\end{tabular}
\caption{Inter-rater reliability for pairs P1 and P2.}
\label{tab:irr}
\end{table}
Table \ref{tab:irr} shows that kappa for argumentation ranges from $0.61 - 0.8$, which generally indicates substantial agreement \cite{Mchugh:12}. Kappa values for specificity and knowledge domain are in the $0.81 - 1$ range which generally indicates almost perfect agreement \cite{Mchugh:12}.
These results show that our proposed annotation scheme can be used to produce reliable annotations of classroom discussion with respect to argumentation, specificity, and knowledge domain.

\begin{table}[t]
\centering
\begin{tabular}{|r|c|c|c|}
\hline
\multicolumn{1}{|l|}{\textbf{Argumentation}}                                              & \textbf{evidence} & \textbf{warrant} & \textbf{claim} \\ \hline
\textbf{evidence}                                                                         & 25                & 5                & 0              \\ \hline
\textbf{warrant}                                                                          & 6                 & 92               & 12             \\ \hline
\textbf{claim}                                                                            & 0                 & 2                & 27             \\ \hline \hline
\multicolumn{1}{|l|}{\textbf{Specificity}}                                                & \textbf{low}      & \textbf{medium}     & \textbf{high}  \\ \hline
\textbf{low}                                                                              & 59                & 5                & 3              \\ \hline
\textbf{medium}                                                                              & 5                 & 25               & 2              \\ \hline
\textbf{high}                                                                             & 1                 & 6                & 63             \\ \hline \hline
\multicolumn{1}{|l|}{\textbf{\begin{tabular}[c]{@{}l@{}}Knowledge\\ Domain\end{tabular}}} & \multicolumn{1}{|c|}{\textbf{\begin{tabular}[c]{@{}l@{}}discipl-\\ inary\end{tabular}}}  & \multicolumn{1}{|c|}{\textbf{\begin{tabular}[c]{@{}l@{}}experi-\\ ential\end{tabular}}}   &                \\ \hline
\textbf{disciplinary}                                                                          & 138               & 1                &                \\ \hline
\textbf{experiential}                                                                            & 0                 & 30               &                \\ \hline
\end{tabular}
\caption{Confusion matrices for argumentation, specificity, and knowledge domain, for annotator pair P1.}
\label{tab:confusion_matrices}
\end{table}

Table \ref{tab:confusion_matrices} shows confusion matrices\footnote{The class distributions for argumentation and specificity labels vary significantly across transcripts, as can be seen in \cite{Lugini:17} and \cite{Godley:17}.}
for annotator pair P1 (we observed similar trends for P2). The argumentation section of the table shows that the largest number of disagreements happens between the claim and warrant labels. One reason may be related to the constraint we impose on warrants - they require the existence of a claim and evidence. If a student tries to provide a warrant for a claim that happened much earlier in the discussion, the annotators might interpret the warrant as new claim.
The specificity section shows relatively few low-high label disagreements as compared to low-med and med-high.
This is also reflected in the quadratic-weighted kappa as low-high disagreements will carry a larger penalty (unweighted kappa is $0.797$).
The main reasons for disagreements over specificity labels come from two of the four specificity elements discussed in Section 3.2: whether an argument move is related to one character or scene, and whether it provides a chain of reasons. With respect to the first of these two elements we observed disagreements in argument moves containing pronouns with an ambiguous reference. Of particular note is the pronoun \textit{it}. If we consider the argument move \textit{``I mean even if you know you have a hatred towards a standard or whatever, you still don't kill it"}, the pronoun \textit{it} clearly refers to something within the move (i.e. the standard) that the student themselves mentioned. In contrast, for argument moves such as \textit{``It did happen"} it might not be clear to what previous move the pronoun refers, therefore creating confusion on whether this specificity element is accomplished.
Regarding specificity element (4) we found that it was easier to determine the presence of a chain of reasons when discourse connectives (e.g. because, therefore) were present in the argument move. The absence of explicit discourse connectives in an argument move might drive annotators to disagree on the presence/absence of a chain of reasons, which is likely to result in a different specificity label.
Additionally, annotators found that shorter turns at talk proved harder to annotate for specificity.
Finally, as we can see from the third section in the table, knowledge domain has the lowest disagreements with only one.

We also \cite{Godley:17} explored the validity of our coding scheme by comparing our annotations of student talk to English Education experts' evaluations (quadratic-weighted kappa of 0.544) of the discussion's quality. Using stepwise regressions, we found that the best model of discussion quality (R-squared of $0.432$) included all three of our coding dimensions: argumentation, specificity, and knowledge domain.

\section{Opportunities and Challenges}
Our annotation scheme introduces opportunities for the educational community to conduct further research on the relationship between features of student talk, student learning, and discussion quality.
Although Chisholm and Godley \shortcite{Chisholm:11} and we found relations between our coding constructs and discussion quality, 
these were small-scale studies based on manual annotations. Once automated classifiers are developed, such relations between talk and learning can be examined at scale. Also, automatic labeling via a standard coding scheme can support the generalization of findings across studies, and potentially lead to
automated tools for teachers and students.

The proposed annotation scheme also introduces NLP opportunities and challenges.
Existing systems for classifying specificity and argumentation have largely been designed to analyze written text rather than spoken discussions.
This is (at least in part) due to a lack of publicly available corpora and schemes for annotating argumentation and specificity in spoken discussions. The development of an annotation scheme explicitly designed for this problem is the first step towards collecting and annotating corpora that can be used by the NLP community to advance the field in this particular area.
Furthermore, in text-based discussions, NLP methods need to tightly couple the discussion with contextual information (i.e., the text under discussion).
For example, an argument move from one of the discussions mentioned in Section 4 stated \textit{``She's saying like free like,  I don't have to be, I don't have to be this salesman's wife anymore, your know? I don't have to play this role anymore."} The use of the term \textit{salesman} shows the presence of specificity element (3) (see Section 3.2) because the text under discussion is indeed \textit{Death of a Salesman}. If the students were discussing another text, the mention of the term \textit{salesman} would not indicate one of the specificity elements, therefore lowering the specificity rating.
Thus, using existing systems is unlikely to yield good performance.  In fact, we previously \cite{Lugini:17} showed that while using an off-the-shelf system for predicting specificity in newspaper articles resulted in low performance when applied to classroom discussions, 
exploiting characteristics of our data could significantly improve performance.
We have similarly evaluated the performance of two existing argument mining systems \cite{Nguyen:18,Niculae:17} on the transcripts described in Section~\ref{sec:data}.
We noticed that since the two systems were trained to classify only claims and premises, they were never able to correctly predict warrants in our transcripts.
Additionally, both systems classified the overwhelming majority of moves as premise, resulting in negative kappa in some cases.
Using our  scheme to create a corpus of classroom discussion data manually annotated for argumentation, specificity, and knowledge domain will support the development of more robust NLP prediction systems.

\section{Conclusions}
In this work we proposed a new annotation scheme for three theoretically-motivated features of student talk in classroom discussion: argumentation, specificity, and knowledge domain.
We demonstrated usage of the scheme by presenting an annotated  excerpt of a classroom discussion. We demonstrated that the scheme can be annotated with high reliability and reported on scheme validity. Finally, we discussed some possible applications and challenges posed by the proposed annotation scheme for both the educational and NLP communities.
We plan to extend our annotation scheme to label information about collaborative relations between different argument moves, and release a corpus annotated with the extended scheme.

\section*{Acknowledgements}
We want to thank Haoran Zhang, Tazin Afrin, and Annika Swallen for their contribution, and all the anonymous reviewers for their helpful suggestions.

This work was supported by the Learning Research and Development Center at the University of Pittsburgh.


\bibliography{bea_2018_annotations}
\bibliographystyle{acl_natbib}

\end{document}